\pdfoutput=1

\documentclass[11pt]{article} 

\usepackage[preprint]{acl}

\usepackage{times}
\usepackage{latexsym}

\usepackage[T1]{fontenc}

\usepackage[utf8]{inputenc}

\usepackage{microtype}

\usepackage{inconsolata}

\usepackage{booktabs}
\usepackage{todonotes}
\usepackage{longtable}
\usepackage{multirow}
\usepackage{siunitx} 
\usepackage{svg}
\usepackage{xcolor}

\newcommand{\nm}[1]{\textcolor{black}{#1}}

\title{Beyond Hate Speech: NLP's Challenges and Opportunities in Uncovering Dehumanizing Language}

\author{Hamidreza Saffari\thanks{\enspace Equal contribution.}\textsuperscript{1}, \space 
  Mohammadamin Shafiei$^{*}$\textsuperscript{2}, \space  
  Hezhao Zhang$^{*}$\textsuperscript{4},\\
  \textbf{Lasana Harris\textsuperscript{3}}, \space
   \textbf{Nafise Sadat Moosavi\textsuperscript{4}} \\
\textsuperscript{1}Politecnico di Milano, 
\textsuperscript{2}University of Milan, 
\textsuperscript{3}University College London, 
\textsuperscript{4}University of Sheffield \\
\texttt{hamidreza.saffari@mail.polimi.it} \\
\texttt{m.shafieiapoorvari@studenti.unimi.it} \\
\texttt{lasana.harris@ucl.ac.uk} \\
\texttt{\{hzhang181, n.s.moosavi\}@sheffield.ac.uk} \\
}

\begin{document}
\maketitle
\begin{abstract}

Dehumanization, i.e., denying human qualities to individuals or groups, is a particularly harmful form of hate speech that can normalize violence against marginalized communities. Despite advances in NLP for detecting general hate speech, approaches to identifying dehumanizing language remain limited due to scarce annotated data and the subtle nature of such expressions. In this work, we systematically evaluate four state-of-the-art large language models (LLMs) — Claude, GPT, Mistral, and Qwen — for dehumanization detection.
Our results show that only one model—Claude—achieves strong performance (over 80\% F$_1$) under an optimized configuration, while others, despite their capabilities, perform only moderately. Performance drops further when distinguishing dehumanization from related hate types such as derogation. We also identify systematic disparities across target groups: models tend to over-predict dehumanization for some identities (e.g., Gay men), while under-identifying it for others (e.g., Refugees). These findings motivate the need for systematic, group-level evaluation when applying pretrained language models to dehumanization detection tasks.

\end{abstract}

\section{Introduction}
Dehumanization, defined as the denial of `humanness' to others \citep{haslam2006dehumanization}, significantly impacts society by fostering conditions that result in extreme and violent behaviors against marginalized groups \citep{KTEILY2022222}. This phenomenon can range from overt derogation, where victims are likened to `dogs' or `monkeys' \citep{hagan2008collective}, to subtler forms, such as denying the capability of experiencing pain to certain individuals \citep{deska2020race}. The identification of dehumanizing language is crucial for understanding and mitigating its effects on collective violence and the manipulation of public perception in conflicts \citep{oberschall1997}.
However, existing hate speech datasets rarely contain sufficient instances of dehumanizing content, and current models struggle to distinguish such language from more benign forms of hate or offense.
Given its uniquely harmful impact and the limited research attention it has received, we believe dehumanization warrants specific focus. Our work is motivated by the need to better understand and identify this extreme form of marginalization, thereby supporting broader efforts in social science and policy aimed at addressing its consequences.

This study evaluates the capabilities of four prominent LLMs — Claude, GPT, Mistral, and Qwen — in accurately identifying dehumanizing language. Through a comprehensive analysis across zero-shot, few-shot, and explainable prompting settings, we assess these models' effectiveness in distinguishing dehumanizing content from other forms of hate speech. Our findings reveal stark performance disparities: in the best configuration, Claude achieves 84.60\% accuracy and 85.82\% F$_1$(dehum.) score in identifying general dehumanization, but performance drops significantly to 78.42\% accuracy when distinguishing dehumanization from other forms of hate speech. Few-shot prompting dramatically improves performance over zero-shot approaches, increasing Claude's accuracy from 54.19\% to 78.42\% in the challenging dehumanization versus hate task. We observe concerning error patterns across target groups: models frequently misclassify other hate types as dehumanization for certain populations (Gay men, Chinese people, and Trans people), while failing to detect genuine dehumanizing language targeting vulnerable groups like Refugees, Immigrants, and South Asians. 
These biases manifest in specific patterns of hate type confusion—models particularly struggle to differentiate dehumanization from derogation, animosity and threatening. This systematic confusion between explicit aggression and the more nuanced denial of humanness raises important questions about LLMs' ability to capture the theoretical distinctions between related forms of harmful language.

\paragraph{Our contributions are:}
    (1) We present the first systematic evaluation of state-of-the-art language models on the task of identifying dehumanizing language, comparing different prompting strategies and labeling criteria.
    (2) We find that while models can perform well on general dehumanization detection, they often conflate it with related hate types such as derogation, indicating a lack of semantic distinction in their predictions.
    (3) We identify systematic disparities in model performance across target groups, revealing patterns of differential sensitivity that raise fairness concerns for content moderation and social analysis.
    (4) We include linguistically motivated baselines to contextualize model performance and assess whether simple heuristics are sufficient for this task.

\section{Related Work}

\nm{Dehumanization has long been a central focus in social science, where researchers have explored its psychological underpinnings and societal consequences \cite{paladino2002differential,haslam2006dehumanization,haslam2008subhuman,haslam2014dehumanization,KTEILY2022222,harris2015dehumanized,leyens2000emotional}. In contrast, computational approaches to dehumanization remain limited, despite the clear potential of natural language processing  to scale such analysis and uncover new patterns in digital discourse.}
\nm{The first notable step toward a computational treatment of dehumanization was made by \newcite{Mendelsohn2020}, who proposed a framework using traditional NLP methods, including word2vec embeddings \cite{mikolov2013efficient} and connotation frames \cite{rashkin-etal-2016-connotation}, to analyze 30 years of New York Times articles. Their study focused on LGBTQ-related terms and measured dehumanization through four conceptual dimensions: Negative Evaluation, Denial of Agency, Moral Disgust, and Vermin Metaphors. While insightful at the aggregate level, their method faces two main limitations: (1) it does not localize specific dehumanizing spans within text, and (2) it is not well-suited for shorter or noisier genres like social media content.}
\nm{Subsequent work has positioned dehumanization within broader psychological constructs. \citet{friedman2021toward} model it as part of moral disengagement, using a small manually annotated dataset to train a SpanBERT model and build a relational knowledge graph. Their data is not publicly available, limiting reproducibility.}

\nm{We use the dataset from \citet{vidgen2020learning} as the basis of our evaluation. It contains 41K social media posts labeled for various hate speech types, including 906 examples of dehumanization, along with target group annotations. This allows us to assess both whether models can distinguish dehumanization from other hate speech and how model behavior varies across different targeted groups.}

\nm{Other recent efforts include \citet{engelmann-etal-2024-dataset}, who present a corpus and a smaller annotated set (918 examples) focused on political and cinematic discourse. While valuable, their dataset lacks target group annotations and frames the task as binary classification, dehumanizing vs. non-dehumanizing, without distinguishing dehumanization from other forms of hate. This limits its utility for analyzing model biases across identity groups or assessing whether models can differentiate between dehumanization and related, but distinct, hate speech.
\citet{10.1145/3711542.3711598} take a lexicon-based approach, identifying dehumanizing content using a curated term list. While effective for flagging explicit language, such approaches are limited in capturing subtle or implied forms of dehumanization, such as metaphor, denial of mental states, or rhetorical framing, which often evade keyword-based detection.
}
\nm{These studies underscore the challenges of modeling dehumanization: it is conceptually diffuse, often implicit, and sparsely represented in existing annotated corpora. Prior approaches have largely depended on task-specific models and small, domain-specific datasets, which limits generalizability and poses barriers for interdisciplinary research.
Recent advances in LLMs offer an opportunity to revisit this problem without relying on supervised training. Pretrained models can perform well in zero- and few-shot settings, potentially enabling more scalable and accessible methods for detecting dehumanization. However, their reliability in this domain remains underexplored, particularly in distinguishing dehumanization from other hate speech and in maintaining consistent performance across target groups.}

\nm{This work contributes to addressing these gaps by systematically evaluating state-of-the-art LLMs for their ability to detect dehumanizing language, with particular attention to model biases and generalization.}

\section{Experimental Setup}
\label{sect:baselines}

\subsection{Dataset}


As mentioned, we use \citet{vidgen2020learning}'s publicly available hate speech dataset for our evaluations. This dataset contains over 41{,}000 adversarially curated entries labeled as \textit{Hate} or \textit{Not Hate}. For entries labeled `Hate', there are secondary labels for the type of hate and the target of hate.  The types of hate include \textit{`Derogation', `Animosity', `Threatening language', `Support for hateful entities',} and \textit{`Dehumanization'.}
For our analysis, we specifically use the \textit{Dehumanization} subset, which consists of 906 entries labeled as \textit{Hate} and further annotated as expressing \textit{Dehumanization}. The inclusion of other hate speech types in this dataset allows us to evaluate the model's ability to distinguish \textit{`Dehumanization'} from various forms of hate speech.
The inclusion of target groups allows us to analyze whether the model's performance on dehumanizing instances varies depending on the targeted group. 
\\
\\
Here are the definitions of the Hate types from \citet{vidgen2020learning}:

\paragraph{Derogation} Content which explicitly attacks, demonizes, demeans or insults a group. This is similar to definitions used in other datasets that include `derogatory', `attacks', or `insults', e.g., \textit{``I absolutely hate hearing <TARGET> talk, they sound so silly''}

\paragraph{Animosity} Content which expresses abuse against a group in an implicit or subtle manner. This resembles `implicit' or `covert' categories in other taxonomies, e.g., \textit{``Are there a lot of <TARGET>? Maybe let's go somewhere else, then?''}

\paragraph{Threatening language} Content which expresses intention to, support for, or encourages inflicting harm on a group, or identified members of the group,
e.g., \textit{``I agree. We need to work together to kill all the <TARGET> off''}

\paragraph{Support for hateful entities} Content which explicitly glorifies, justifies or supports hateful actions, events, organizations, tropes, and individuals (collectively, ‘entities’), e.g., \textit{``I wish I was literally in nazi germany in 1937!''}

\paragraph{Dehumanization} Content which ``perceives or treats people as less than human''. It often involves describing groups using terms like leeches, cockroaches, insects, germs, or rats. For example: \textit{``I got attacked by a stupid <TARGET> rat, he took my mobile phone''}




\subsection{Evaluation Subsets}

We devised two evaluation subsets from this dataset to evaluate dehumanization:


\paragraph{General Dehumanization:} This set contains all 906 \textit{dehumanization} instances, which may have different targeted groups, along with 906 randomly selected instances from the dataset. The randomly selected examples contain 414 instances of hate speech and 492 non-hate speech labels. This balanced subset evaluates models' ability to to recognize dehumanization across diverse groups while also measuring their fundamental capability to distinguish hate from non-hate content. \footnote{The frequency distribution of target groups across our evaluation subsets can be found in Appendix \ref{app:target_frequencies}.}

\paragraph{Dehumanization vs. Hate:} It consists of 906 instances of \textit{dehumanization} as well as 906 randomly selected instances from other hate speech labels (\textit{Derogation}: 652, \textit{Animosity}: 209, \textit{Threatening}: 36, \textit{Support}: 9), testing the model's ability to distinguish between \textit{dehumanization} and other forms of hate speech.

\subsection{Baselines}

\nm{To better understand how specific linguistic signals relate to dehumanizing language, we incorporate four interpretable components from \citet{Mendelsohn2020}'s framework as baselines. These components reflect common patterns in dehumanizing discourse: expressing negativity toward a group, portraying them as lacking agency, evoking moral disgust, or comparing them to vermin. We adapt these components to assess how well each can identify dehumanization in our evaluation data. The full methodological details for these components can be found in \citet{Mendelsohn2020}, we summarize their core logic here for clarity. For comparison with more recent approaches, we also evaluate a RoBERTa-based classifier (see Appendix \ref{app:traditional_comparison} for details).
} 

\begin{table*}[!htb]
\centering
\footnotesize
\resizebox{\textwidth}{!}{

\begin{tabular}{@{}p{2.5cm}p{4.5cm}p{8.5cm}@{}}
\toprule
\textbf{Prompt Type} & \textbf{Label Output} & \textbf{Key Prompt Instructions} \\
\midrule

\textbf{Zero-shot} & Binary (True/False) for each target group &
Identify target groups in the text. Decide whether each target is dehumanized. Respond in JSON format:
\texttt{\{ ``Targets'': [...], ``Dehumanization'': [[target1, true/false], ...] \}} \\

\midrule

\textbf{Few-shot} & Blatant / Subtle / None for each target group &
Given labeled examples, identify target groups and classify each as ``Blatant'', ``Subtle'', or ``None''. Use the format:
\texttt{[ \{ ``Target'': ``...'', ``Dehumanization'': ``Blatant''/``Subtle''/``None'' \}, ... ]} \\

\midrule

\textbf{Explainable} & Blatant / Subtle / None + Explanation &
Same as few-shot, but provide a short explanation for each label:
\texttt{[ \{ ``Target'': ``...'', ``Dehumanization'': ``...'', `Explanation'': ``...'' \}, ... ]} \\
\bottomrule
\end{tabular}
}
\caption{Summary of prompt instructions and expected output formats. }
\label{tab:prompt_summary}
\end{table*}

\paragraph{Negative Evaluation}

\nm{This baseline measures how negatively a text describes a target group. First, we estimate the overall tone of the text using the NRC VAD lexicon \cite{mohammad2018obtaining}, which assigns a valence score to words, ranging from 0 (very negative) to 1 (very positive). We average these scores across the text.
Next, to capture how the text describes the target group specifically, we use the connotation frames lexicon \cite{rashkin2015connotation}, which indicates whether verbs imply positive or negative sentiment toward their subjects. If a text has both a low average valence (below 0.5) and a negative connotation score toward the group, we classify it as expressing negative evaluation.}



\paragraph{Denial of Agency}
\nm{Dehumanizing language often portrays groups as lacking autonomy or control. To detect this, we use a verb lexicon from \citet{sap2017connotation} that rates verbs based on whether they suggest high or low agency. For example, verbs like decide or lead imply agency, while suffer or obey do not. We calculate how frequently low-agency verbs are used in a text. If they dominate, we consider the text to deny agency to the group. If no verbs from the lexicon are found, we mark the text as `neutral'.}

\paragraph{Moral Disgust}
\nm{Another common feature of dehumanizing language is associating the target group with moral wrongdoing. To capture this, we use a moral disgust lexicon from \citet{graham2009liberals}, which includes words like sin, disgust, and pervert. We follow \citet{Mendelsohn2020} in computing a vector representation of moral disgust by averaging the word embeddings of these terms. We then compare this vector to the embedding of the input text using cosine similarity. A higher similarity suggests a stronger association with moral disgust.}



\paragraph{Vermin Metaphors}
\nm{Dehumanization often involves metaphorically comparing people to pests or vermin. Following \citet{Mendelsohn2020}, we construct a vector representation based on terms like vermin, rat, cockroach, and bedbug. We then compute the similarity between this vector and the input text’s embedding. A high similarity indicates the presence of vermin metaphors, which are strongly linked to dehumanizing intent.}

\subsection{Models}
We evaluated four state-of-the-art LLMs: Claude-3-7-Sonnet-20250219 \cite{claude3.7sonnet2025}, GPT-4.1-mini-2025-04-14 \cite{openai2024gpt4technicalreport}, Mistral-7B-Instruct-v0.2 \cite{jiang2023mistral7b}, and Qwen2.5-7B-Instruct \cite{qwen2025qwen25technicalreport}. For accessing these models, we used different APIs based on availability: the Anthropic API in batch processing mode for Claude, the OpenAI API in batch processing mode for GPT, and the Hugging Face Inference API for Mistral and Qwen. For consistency and to eliminate randomness, we set the temperature to zero across all evaluations. Throughout our experiments, we refer to these models by their shortened names: Claude, GPT, Mistral, and Qwen.




The effectiveness of state-of-the-art language models often depends on how tasks are framed through prompts. We evaluate three prompting strategies, zero-shot, few-shot, and explainable, each guiding the model to produce different types of outputs. Table~\ref{tab:prompt_summary} outlines their key differences.\footnote{Full prompt templates are provided in Appendix \ref{app_prompt_template}.}


\begin{table*}[!htb]
\centering
\small
\sisetup{table-format=2.2}
\begin{tabular}{lllcS[table-format=2.2]S[table-format=2.2]S[table-format=2.2]|S[table-format=2.2]S[table-format=2.2]S[table-format=2.2]}
\toprule
\textbf{Model} & \textbf{Prompt} & \textbf{Label} & & 
\multicolumn{3}{c|}{\textbf{General Dehumanization}} & 
\multicolumn{3}{c}{\textbf{Dehumanization vs Hate}} \\
\cmidrule(lr){5-7} \cmidrule(lr){8-10}
 & & \textbf{Criterion} & & 
\textbf{F$_1$(other)} & \textbf{F$_1$(dehum.)} & \textbf{Acc.} &
\textbf{F$_1$(hate)} & \textbf{F$_1$(dehum.)} & \textbf{Acc.} \\
\midrule

\multirow{5}{*}{GPT}
 & Zero-shot & Binary & & 37.73 & 58.61 & 50.28 & 25.91 & 61.70 & 49.50 \\
 & Few-shot & Blatant only & & 51.65 & 48.23 & 50.00 & 48.95 & 49.51 & 49.23 \\
 & Explainable & Blatant only & & 52.22 & 45.08 & 48.90 & 50.56 & 47.00 & 48.84 \\
 & Few-shot & Blatant+Subtle & & 30.05 & 60.02 & 49.12 & 19.48 & 64.03 & 50.28 \\
 & Explainable & Blatant+Subtle & & 29.64 & 59.88 & 48.90 & 16.97 & 64.00 & 49.78 \\
\midrule

\multirow{5}{*}{Qwen}
 & Zero-shot & Binary & & 51.23 & 71.42 & 63.96 & 31.42 & 66.72 & 55.19 \\
 & Few-shot & Blatant only & & 73.91 & 71.34 & 72.68 & 68.57 & 68.29 & 68.43 \\
 & Explainable & Blatant only & & 71.97 & 67.81 & 70.03 & 65.19 & 63.46 & 64.35 \\
 & Few-shot & Blatant+Subtle & & 50.78 & 73.25 & 65.34 & 29.24 & 68.55 & 56.46 \\
 & Explainable & Blatant+Subtle & & 49.53 & 72.70 & 64.57 & 27.83 & 67.68 & 55.35 \\
\midrule

\multirow{5}{*}{Mistral}
 & Zero-shot & Binary & & 58.33 & 73.28 & 67.44 & 38.49 & 67.93 & 57.84 \\
 & Few-shot & Blatant only & & 53.69 & 55.65 & 54.69 & 53.05 & 55.28 & 54.19 \\
 & Explainable & Blatant only & & 60.57 & 54.83 & 57.89 & 60.10 & 54.16 & 57.33 \\
 & Few-shot & Blatant+Subtle & & 47.19 & 63.59 & 56.90 & 42.97 & 62.27 & 54.58 \\
 & Explainable & Blatant+Subtle & & 50.18 & 68.67 & 61.53 & 43.49 & 66.70 & 58.09 \\
\midrule

\multirow{5}{*}{Claude}
 & Zero-shot & Binary & & 56.90 & 75.57 & 68.82 & 20.50 & 67.83 & 54.19 \\
 & Few-shot & Blatant only & & 81.74 & 84.67 & 83.33 & 75.05 & 80.99 & 78.42 \\
 & Explainable & Blatant only & & 83.16 & 85.82 & 84.60 & 72.49 & 80.06 & 76.88 \\
 & Few-shot & Blatant+Subtle & & 53.14 & 75.12 & 67.49 & 17.06 & 68.04 & 53.86 \\
 & Explainable & Blatant+Subtle & & 50.12 & 74.53 & 66.28 & 14.13 & 67.68 & 53.04 \\
\midrule

\multicolumn{3}{l}{Negative Evaluation} & & 66.37 & 5.47 & 50.39 & 65.87 & 5.41 & 49.83 \\
\multicolumn{3}{l}{Denial of Agency} & & 62.18 & 18.01 & 48.23 & 63.33 & 18.34 & 49.39 \\
\multicolumn{3}{l}{Moral Disgust} & & 53.60 & 53.13 & 53.37 & 55.32 & 54.27 & 54.80 \\
\multicolumn{3}{l}{Vermin Metaphors} & & 54.63 & 52.38 & 53.53 & 56.24 & 53.85 & 55.08 \\
\multicolumn{3}{l}{Combination} & & 66.74 & 0.66 & 50.17 & 66.69 & 0.66 & 50.11 \\
\bottomrule
\end{tabular}
\caption{Comparison of model performance across prompt types and labeling criteria, evaluated on general dehumanization and dehumanization versus hate speech.}
\label{tab:results_merged}
\end{table*}

\paragraph{Zero-shot:} The prompt consists of the phrase \textit{``Identify target groups and decide if they're dehumanized''}. This scheme assesses the model's preexisting knowledge about dehumanization. 

\paragraph{Few-shot:}
We enhance the model's exposure by incorporating five randomly selected instances of dehumanization targeting frequent targets. In the few-shot setting, the model goes further by classifying dehumanization within texts as either \textit{`blatant'} or \textit{`subtle'}. The included few-shot examples with dehumanizing language are labeled as \textit{`blatant'}. 


\paragraph{Explainable Prompting:} 
Building on the few-shot setting, this approach further requires the model to provide explanations for its decisions.

\section{Results}
\nm{Table~\ref{tab:results_merged} presents model performance across \texttt{zero-shot}, \texttt{few-shot}, and \texttt{explainable} prompting strategies, evaluated under two labeling criteria: (1) considering only \textit{blatant} cases as dehumanization, and (2) treating both \textit{blatant} and \textit{subtle} cases as positive instances. We also include four linguistically motivated baselines from \citet{Mendelsohn2020}. The \textit{Combination} heuristic flags a text only if all four features are present.
The \texttt{zero-shot} setting produces binary dehumanization labels, while the \texttt{few-shot} and \texttt{explainable} settings support multi-class outputs that distinguish between subtle and blatant forms of dehumanization. This allows us to analyze model performance under both strict and inclusive interpretations of dehumanization.}
In all settings, we evaluate each identified target against the relevant criteria. If any target meets the criteria, the text is classified as dehumanizing.

\nm{\paragraph{General Dehumanization.}
Models perform best in the general dehumanization task, where they distinguish dehumanizing from neutral or unrelated content. Claude achieves the highest F$_1$(dehum.) of 85.82\% in the \texttt{explainable} setting under the \textit{blatant-only} criterion, with other models such as Qwen and Mistral also reaching scores above 70\%. When including \textit{subtle} cases, performance is more varied: most models maintain or slightly improve their F$_1$(dehum.), suggesting that subtle examples provide additional positive signal. The exception is Claude, which shows a noticeable drop when \textit{subtle} cases are included—falling from 85.82\% to 74.53\%—indicating higher sensitivity to the ambiguity of these instances.
}

\paragraph{Dehumanization vs. Hate.} 
\nm{
 Distinguishing dehumanization from other types of hate speech proves substantially more difficult. Accuracy across most models remains close to the random baseline (50–58\%), and F$_1$(dehum.) scores generally cluster in the 60\%s. Claude is the notable exception: under the \texttt{few-shot} setting with the \textit{blatant-only} criterion, it achieves an F$_1$(dehum.) of 80.99\% and an accuracy of 78.42\%, outperforming all other configurations by a wide margin. This suggests that Claude is particularly effective at detecting overtly dehumanizing content, though its performance declines when \textit{subtle} cases are included. Other models, including GPT, Qwen, and Mistral, either benefit slightly or maintain stable performance when \textit{subtle} instances are treated as dehumanization, reflecting more flexible, though less precise, behavior.
}

\paragraph{Lexical Baselines.}
\nm{The feature-based baselines show limited effectiveness. \textit{Negative Evaluation} and \textit{Denial of Agency} fail to identify dehumanization reliably. \textit{Moral Disgust} and \textit{Vermin Metaphors} yield slightly more balanced performance, though still substantially below model-based approaches. The \textit{Combination} baseline, which requires all four cues to be present, performs worst overall, indicating that dehumanization is potentially signaled by only a subset of these features.}

\paragraph{Summary.}
\nm{ Overall, LLMs demonstrate strong potential for identifying dehumanizing language in general contexts, particularly when contrasted with neutral content. However, distinguishing dehumanization from other forms of hate speech remains challenging for most models, with performance typically only modestly above chance. Claude is the exception: under the \texttt{few-shot} setting with the \textit{blatant-only} criterion, it achieves markedly higher performance, suggesting that it is particularly effective at detecting overt dehumanizing cues. Our results show that prompt format and label interpretation (blatant vs. subtle) significantly influence outcomes, revealing both the flexibility and fragility of current models when applied to nuanced classification tasks.}

\section{Target Group-Level Performance}

\begin{figure}[!htb]
    \centering
    \includesvg[width=1\linewidth]{figures/f1_scores_by_model.svg}
\caption{F$_1$(dehum.) scores of each model, evaluated under their best-performing configurations, on the 10 most frequent target groups.}
    \label{fig:target_error}
\end{figure}

To investigate how equitably models treat different social groups, we analyze model behavior across target groups in the \texttt{Dehumanization vs. Hate} subset. Figure~\ref{fig:target_error} shows the F$_1$ score for dehumanization for each model using its best-performing configuration (i.e., the optimal combination of prompt type and label criterion) on the 10 most frequent targets in the evaluation set.

We observe that models differ substantially in their consistency across groups. Claude demonstrates relatively uniform performance across most groups, with slightly lower F$_1$ scores for Trans people and Women. In contrast, the other models (Qwen, Mistral, and GPT) show wider disparities: performance drops significantly for Women, Muslims, and Trans people. These findings highlight potential fairness concerns, especially in lower-capacity or less robust models.

\begin{figure}[!htb]
    \centering
    \includesvg[width=1\linewidth]{figures/mistral_claude_false_negatives_comparison.svg}
    \caption{Recognition blindness of Claude and Mistral.}
    \label{fig:claude_false_neg}
\end{figure}

\begin{figure}[!htb]
    \centering
    \includesvg[width=1\linewidth]{figures/mistral_claude_false_positives_comparison.svg}
    \caption{Over-sensitivity of Claude and Mistral.}
    \label{fig:claude_false_pos}
\end{figure}

To better understand the nature of these disparities, we further analyze two types of errors: \textbf{recognition blindness}, where the model fails to detect dehumanization (false negatives ratio), and \textbf{over-sensitivity}, where it incorrectly predicts dehumanization on non-dehumanizing hate speech (false positives ratio). These errors are reported as ratios, calculated by dividing each by the total number of instances for the corresponding target. This normalization allows us to account for differences in target group frequencies and enables fairer comparison across groups. We focus on Claude and Mistral, the two best-performing models, and plot the union of their top 10 highest-error target groups in each category.
Figure~\ref{fig:claude_false_neg} presents recognition blindness. Mistral shows higher false negative ratios for groups such as Refugees and South Asians, even under its best-performing configuration. Claude’s recognition blindness is higher for groups like Lesbians and Immigrants.
Figure~\ref{fig:claude_false_pos} shows over-sensitivity. 
We observe that false positive rates are substantially higher than false negative rates across most target groups. Mistral, in particular, shows pronounced over-sensitivity for Gay men, Chinese people, and Trans people. Claude also exhibits elevated false positives for some of these groups, though at consistently lower levels overall. \footnote{These errors are statistically significant; see Appendix~\ref{app:significance}.}

Taken together, these results show that while both models exhibit group-specific disparities, the nature of errors differs. For some targets (e.g., Gay men), models are prone to over-sensitivity, whereas for others (e.g., Refugees or Lesbians), recognition failures are more common. Overall, over-sensitivity is the dominant error mode, raising concerns for deploying dehumanization detection systems, particularly in settings where false positives carry social or policy implications.

We further examined concrete examples of model failures to better understand their misclassification behavior. We observed two key patterns: (1) models often misclassify highly charged language (e.g., ``STFU'' or threats of violence) and stigmatizing terms such as ``illegals'' as dehumanizing, even when the underlying message does not meet that threshold; and (2) they frequently overlook subtle or metaphorical dehumanization, especially when animal comparisons or the denial of agency is implied rather than explicitly stated. Table~\ref{tab:error_examples} and Table~\ref{tab:error_examples_unrecognized_dehuman} provide some representative examples.

\begin{table*}[!htb]
    \centering
    \small
    \resizebox{\textwidth}{!}{%
    \begin{tabular}{p{9cm}llll} \toprule
    {Text} & {Label} & {Zero} & {Few} & {Exp.} \\ \midrule
    Why don't <XX> just STFU and stay into the kitchen & Derogation & Dehuman & Blatant & Blatant \\ \midrule
    Next time I see a <ZZ> person, I'm going to b e a t them up & Threatening & Dehuman & Blatant & Blatant \\ \midrule
    Someone who had strong year and made over 75k in 2018 and who may be struggling currently gets zero but illegals and refugees get \$350 mil & Animosity & Dehuman & Blatant & Subtle \\ 
    \bottomrule
    \end{tabular}
    }
    \caption{Examples from Claude outputs when the model \textbf{incorrectly classifies other types of hate speech as \textit{dehumanization}.} Some parts of input texts are masked out to reduce their toxicity. }
    \label{tab:error_examples}
\end{table*}

\begin{table*}[!htb]
    \centering
    \small
    \begin{tabular}{p{9cm}ccc} \toprule
    {Text} & {Zero} & {Few} & {Exp.} \\ \midrule
    Animal rights activists back the fuck off, <XX> is the same shit anyway  & D & N & D \\
    \midrule
    Monkeys freak the shit out of me they're so <XX> & D & D & N \\
    \midrule
    What's smarter, a dog or a <YY>… I know, just an easy one to get started. Dogs are obviously smarter. & N & N & D \\
    \bottomrule
    \end{tabular}
    \caption{Examples where \textbf{\textit{dehumanization} is not detected by Claude or Mistral.} Some parts of the input texts are masked to reduce toxicity. D: Correctly detected as \textit{dehumanization}, ND: Not Detected.}
    \label{tab:error_examples_unrecognized_dehuman}
\end{table*}

\begin{figure}[!htb]
    \centering
    \includesvg[width=0.8\linewidth]{figures/combined_best_sankey.svg}
    \caption{Distribution of {other hate types that were misclassified as \textit{dehumanization}}.}
    \label{fig:mistaken_hate}
\end{figure}

\begin{figure}[!htb]
    \centering
    \includesvg[width=0.8\linewidth]{figures/models_to_misclassifications_sankey-few.svg}
    \caption{Misclassification of \textit{dehumanization} as other hate types by Claude and GPT in the fine-grained classification task.}
    \label{fig:models_mis}
\end{figure}

\section{Misclassification Patterns Across Hate Types}
Figure~\ref{fig:mistaken_hate} shows the distribution of hate speech types that Claude and Mistral, under their best-performing configurations, misclassify as \textit{dehumanization} in the \texttt{Dehumanization vs. Hate} subset. The majority of errors arise from \textit{derogation}, followed by \textit{animosity} and \textit{threatening}, suggesting that models often conflate general hostility, subtle abuse, or explicit threats with the more specific construct of dehumanization. 

To further examine how models confuse dehumanization with other hate types, we conduct a fine-grained hate type classification experiment using the \texttt{Dehumanization vs. Hate} subset. Unlike the earlier binary task, this setting requires models to assign each input to one of several predefined hate types, including \textit{dehumanization}, \textit{derogation}, \textit{animosity}, \textit{threatening}, and \textit{support}. \footnote{This classification task is detailed in Appendix \ref{app:fine_grained}.}
We focus on two models: Claude (best-performing) and GPT (worst-performing). Both models are evaluated using a \texttt{few-shot} setting where an example of each hate label is provided.
Figure~\ref{fig:models_mis} shows how each model misclassifies true \textit{dehumanization} examples in this setting. Both Claude and GPT most frequently confuse dehumanization with \textit{derogation}, suggesting substantial semantic overlap.  GPT exhibits broader confusion across hate types but remains most prone to overgeneralizing \textit{dehumanization} as \textit{derogation}.
To better understand overall confusion patterns, Figures~\ref{fig:claude_cm} and~\ref{fig:gpt_cm} show the full confusion matrices for Claude and GPT. Claude achieves relatively strong separation between classes, especially for \textit{support} and \textit{threatening}, but continues to mislabel a notable portion of \textit{dehumanization} instances as derogation. GPT performs less consistently, with heavier confusion between \textit{dehumanization}, \textit{derogation}, and \textit{animosity}. It misclassifies about one-third of \textit{dehumanization} cases as \textit{derogation}, and also struggles more on animosity.
These results suggest that even high-performing models have difficulty distinguishing \textit{dehumanization} from closely related hate types.

\begin{figure}[!htb]
    \centering
    \includesvg[width=0.75\linewidth]{figures/cm_Claude_few.svg}
    \caption{Confusion Matrix of Calude.}
    \label{fig:claude_cm}
\end{figure}

\begin{figure}[!htb]
    \centering
    \includesvg[width=0.75\linewidth]{figures/cm_gpt_few.svg}
    \caption{Confusion Matrix of GPT.}
    \label{fig:gpt_cm}
\end{figure}

\section{Conclusion}
Dehumanizing language plays a distinct and harmful role within hate speech, often serving to justify violence or exclusion against marginalized groups. This paper examined the capabilities of state-of-the-art LLMs to detect dehumanization in both binary and fine-grained classification settings, without task-specific training.
Our findings show that general-purpose models demonstrate promising performance. However, they consistently struggle to distinguish dehumanization from related hate types, particularly derogation. We also identify disparities in model behavior across social groups, with some targets more likely to trigger over-sensitive predictions.
These results show both the promise and limitations of using pretrained language models for dehumanization detection. While such models offer scalable analysis without task-specific training, their inconsistent behavior across target groups raises concerns about fairness and reliability. 

\section{Limitations}

Our study has several limitations worth noting. We relied exclusively on \citet{vidgen2020learning}'s dataset, which contains limited dehumanization instances and may not represent naturally occurring patterns across platforms. Our evaluation approach of prompting rather than fine-tuning LLMs may not reflect optimal model performance, and standard metrics like accuracy and F1 do not fully capture the social harm dimensions of misclassifications. Deploying these models in real-world content moderation risks reinforcing biases, as our analysis revealed target-specific performance disparities that could lead to disproportionate content flagging or oversight. Additionally, such systems might inadvertently suppress legitimate discourse from marginalized groups using reclaimed terminology. Finally, our findings may not generalize to non-English contexts or cultures where dehumanization takes different linguistic forms. Addressing these limitations requires collaboration between NLP researchers, social scientists, and affected communities.

\bibliography{custom}
\appendix

\label{sec:appendix}
\appendix

\section{Prompt Templates and Examples}
\label{app_prompt_template}

This section presents the exact prompt templates used in our experiments and example sets used for few-shot learning.

\subsection{Prompt Templates}

We used three distinct prompting strategies in our experiments. Below are the exact templates for each approach:

\paragraph{Zero-shot Prompt}
\noindent \texttt{You are an expert in psychology.$\backslash$nText: \{text\}.$\backslash$nFirst, Identify target groups in the text$\backslash$nDecide whether each target is dehumanised by the author.$\backslash$nOnly response in the example output format that can be parsed by json:$\backslash$n\{\{$\backslash$n    "Targets": ["group\_name1", "group\_name2", "..."],$\backslash$n    "Dehumanisation": [$\backslash$n        ["target1", true],$\backslash$n        ["target2", false], $\backslash$n        ["...", true]$\backslash$n    ]$\backslash$n\}\}$\backslash$n}

\paragraph{Few-shot Prompt}
\noindent \texttt{You are an expert in psychology and NLP.Examples of dehumanization:$\backslash$n\{examples\}$\backslash$n$\backslash$nAnalyze the following text and identify all the groups targeted in it. For each group, assess whether they are subjected to any form of dehumanization, be it blatant, subtle, or none at all.$\backslash$n$\backslash$nText: $\backslash$n\{text\}$\backslash$n$\backslash$nMake sure target name exist and exactly the same in given text. Only response in the example output json format:$\backslash$n [{{$\backslash$n"target\_name":target name in text, "dehumanize": "None" or "Subtle" or "Blatant"$\backslash$n],...}}]$\backslash$n}

\paragraph{Explainable Prompt}
\noindent \texttt{You are an expert in psychology and NLP.Examples of dehumanization:$\backslash$n\{examples\}$\backslash$n$\backslash$nAnalyze the following text and identify all the groups targeted in it. For each group, assess whether they are subjected to any form of dehumanization, be it blatant, subtle, or none at all.$\backslash$n$\backslash$nText: $\backslash$n\{text\}$\backslash$n$\backslash$nMake sure target name exist and exactly the same in given text. Only response in the example output json format:$\backslash$n [{{$\backslash$n"target\_name":target name in text, "dehumanize": "None" or "Subtle" or "Blatant", "explanation":explanation$\backslash$n],...}}]$\backslash$n}

\subsection{Few-Shot Example Sets}

To evaluate the impact of example selection on model performance, we compiled two distinct sets of five dehumanizing examples. Table \ref{tab:example_sets} presents these example sets, which were chosen to represent various forms of dehumanization targeting different groups.

\begin{table*}[!htb]
    \centering
    \small
    \begin{tabular}{cl}
        \toprule
        \textbf{Example Set} & \textbf{Examples} \\
        \midrule
        Set 1 & \begin{tabular}[m]{@{}l@{}}
            1. ``camel jockeys are the ugliest insects on the planet!'' \\
            2. ``Saudi and Israel, the savage inhabitants of God's monkey house. They are \\
            \hspace{1em} screeching and flinging their faeces at each other again, it never ends.'' \\
            3. ``No more f\#\#\#ing terrorists: Denmark to cull entire Muslim population after \\
            \hspace{1em} terrorist mutation spreads within humans.'' \\
            4. ``black women are vicious animals and should be banned from coming to schools.'' \\
            5. ``asian women r subhummmmmman.'' \\
        \end{tabular} \\
        \midrule
        Set 2 & \begin{tabular}[m]{@{}l@{}}
            1. ``I do think all black ppl r dumb. U know, they would never question a vaccine \\
            \hspace{1em} if it's being given by a black person to a black person, so we should get some \\
            \hspace{1em} black doctors on board. Lol smdh. Then, proceed to give them the vaccine 1st, \\
            \hspace{1em} and treat them like guinea pigs.'' \\
            2. ``These people that are coming are not immigrants, asylum seekers or anything else. \\
            \hspace{1em} They are invaders, pure and simple coming to take over our nation.'' \\
            3. ``The 'jew' tick cries out in pain as you pluck it from your hairy ballsack!'' \\
            4. ``Fleeing your country like refugee rats from (and on!) a sinking ship so that you \\
            \hspace{1em} can live off the taxpayer's money in Europe is not a solution.'' \\
            5. ``All women are not to be trusted, they are snakes, vicious vipers trying to trick their men.'' \\
        \end{tabular} \\
        \bottomrule
    \end{tabular}
    \caption{Examples included in each example set used for evaluating models on dehumanization detection.}
    \label{tab:example_sets}
\end{table*}

\section{Additional Performance Analysis}
\label{app:perf_analysis}

This section provides further details on model performance across different configurations and example sets.

\subsection{Impact of Example Sets on Performance}

Table \ref{tab:results_two_example_sets} compares the performance of Claude and GPT models across both example sets in all prompting configurations. The results demonstrate that example selection impacts model performance, with Set 1 yielding better results for Claude on both evaluation subsets.

\begin{table*}[!htb]
    \centering
    \small
    \begin{tabular}{cllccc|ccc} 
    \toprule
     & & & \multicolumn{3}{c|}{General Dehumanization} & \multicolumn{3}{c}{Dehumanization vs Hate} \\
     Example Set & Model & Prompt & F$_1$(other) & F$_1$(dehum.) & Acc. & F$_1$(hate) & F$_1$(dehum.) & Acc. \\ \midrule
     \multirow{12}{*}{Set 1} 
     & \multirow{6}{*}{Claude} 
         & Zero-shot (Binary) & 56.90 & 75.57 & 68.82 & 20.50 & 67.83 & 54.19 \\
         & & Few-shot (Blatant) & 81.74 & 84.67 & 83.33 & 75.05 & 80.99 & 78.42 \\
         & & Explainable (Blatant) & 83.16 & 85.82 & 84.60 & 72.49 & 80.06 & 76.88 \\
         & & Zero-shot (Binary) & 56.90 & 75.57 & 68.82 & 20.50 & 67.83 & 54.19 \\
         & & Few-shot (Bl.+Subtle) & 53.14 & 75.12 & 67.49 & 17.06 & 68.04 & 53.86 \\
         & & Explainable (Bl.+Subtle) & 50.12 & 74.53 & 66.28 & 14.13 & 67.68 & 53.04 \\ \cmidrule{2-9}
     & \multirow{6}{*}{GPT} 
         & Zero-shot (Binary) & 37.73 & 58.61 & 50.28 & 25.91 & 61.70 & 49.50 \\
         & & Few-shot (Blatant) & 51.65 & 48.23 & 50.00 & 48.95 & 49.51 & 49.23 \\
         & & Explainable (Blatant) & 52.22 & 45.08 & 48.90 & 50.56 & 47.00 & 48.84 \\
         & & Zero-shot (Binary) & 37.73 & 58.61 & 50.28 & 25.91 & 61.70 & 49.50 \\
         & & Few-shot (Bl.+Subtle) & 30.05 & 60.02 & 49.12 & 19.48 & 64.03 & 50.28 \\
         & & Explainable (Bl.+Subtle) & 29.64 & 59.88 & 48.90 & 16.97 & 64.00 & 49.78 \\ \midrule
     \multirow{12}{*}{Set 2}
     & \multirow{6}{*}{Claude} 
         & Zero-shot (Binary) & 57.25 & 75.60 & 68.93 & 21.52 & 67.99 & 54.53 \\
         & & Few-shot (Blatant) & 77.37 & 83.07 & 80.63 & 62.67 & 76.91 & 71.47 \\
         & & Explainable (Blatant) & 78.03 & 83.37 & 81.07 & 60.66 & 76.23 & 70.36 \\
         & & Zero-shot (Binary) & 57.25 & 75.60 & 68.93 & 21.52 & 67.99 & 54.53 \\
         & & Few-shot (Bl.+Subtle) & 46.09 & 73.68 & 64.62 & 10.34 & 67.37 & 52.15 \\
         & & Explainable (Bl.+Subtle) & 45.81 & 73.44 & 64.35 & 8.56 & 67.14 & 51.66 \\ \cmidrule{2-9}
     & \multirow{6}{*}{GPT} 
         & Zero-shot (Binary) & 37.26 & 58.44 & 50.00 & 25.93 & 61.76 & 49.56 \\
         & & Few-shot (Blatant) & 45.75 & 52.28 & 49.23 & 42.49 & 55.69 & 49.94 \\
         & & Explainable (Blatant) & 46.73 & 49.76 & 48.29 & 45.77 & 53.91 & 50.17 \\
         & & Zero-shot (Binary) & 37.26 & 58.44 & 50.00 & 25.93 & 61.76 & 49.56 \\
         & & Few-shot (Bl.+Subtle) & 23.65 & 60.97 & 48.34 & 12.67 & 64.76 & 49.78 \\
         & & Explainable (Bl.+Subtle) & 24.39 & 61.15 & 48.68 & 12.86 & 64.83 & 49.89 \\
    \bottomrule
    \end{tabular}
    \caption{Comparison of identifying dehumanizing language for Claude and GPT models using two different example sets, showing performance on \texttt{General Dehumanization} and \texttt{Dehumanization vs. Hate speech} subsets with different labeling criteria.}
    \label{tab:results_two_example_sets}
\end{table*}

\section{Error Analysis}
\label{app:error_analysis}

This section provides additional details on the error patterns observed in our models, with particular attention to misclassification between different hate types.

\subsection{Misclassification Patterns}

Figures \ref{fig:exp_misclass}, \ref{fig:few_misclass}, and \ref{fig:zero_misclass} show the distribution of hate types misclassified as dehumanization across different prompting strategies. These visualizations reveal consistent patterns of confusion, particularly between derogation and dehumanization, across all models and prompting approaches.

\begin{figure}[!htb]
    \centering
    \includesvg[width=1\linewidth]{figures/combined_exp_sankey.svg}
    \caption{Distribution of \textbf{other hate types that were misclassified as `Dehumanization'} in the \texttt{Dehumanization vs. Hate} subset for each model under the \texttt{Explainable Prompt}.}
    \label{fig:exp_misclass}
\end{figure}

\begin{figure}[!htb]
    \centering
    \includesvg[width=1\linewidth]{figures/combined_few_shot_sankey.svg}
    \caption{Distribution of \textbf{other hate types that were misclassified as `Dehumanization'} in the \texttt{Dehumanization vs. Hate} subset for each model under the \texttt{Few-shot Prompt}.}
    \label{fig:few_misclass}
\end{figure}

\begin{figure}[!htb]
    \centering
    \includesvg[width=1\linewidth]{figures/combined_zero_shot_sankey.svg}
    \caption{Distribution of \textbf{other hate types that were misclassified as `Dehumanization'} in the \texttt{Dehumanization vs. Hate} subset for each model under the \texttt{Zero-shot Prompt}.}
    \label{fig:zero_misclass}
\end{figure}

\subsection{Detailed Misclassification Analysis}

Table \ref{tab:misclass_exp} presents a detailed breakdown of misclassification rates by hate type for Claude and Mistral. This quantitative analysis reveals that while Claude exhibits lower overall misclassification, both models struggle most with distinguishing derogation from dehumanization.

\begin{table*}[!h]
\centering
\renewcommand{\arraystretch}{1.2}
\begin{tabular}{lcc|cc}
\toprule
\multirow{2}{*}{Hate Type} & \multicolumn{2}{c|}{Claude} & \multicolumn{2}{c}{Mistral} \\
& Misclass./Total & Ratio (\%) & Misclass./Total & Ratio (\%) \\
\midrule
Derogation   & 283/652 & 43.4 & 523/652 & 80.2 \\
Animosity    & 18/209 & 8.6 & 108/209 & 51.6 \\
Threatening  & 17/36   & 47.2 & 33/36   & 91.6 \\
Support      & 0/9     & 0 & 3/9     & 33.3 \\
\bottomrule
\end{tabular}
\caption{Hate types that were misclassified as \textit{dehumanization} under each model's best-performing configuration.}
\label{tab:misclass_exp}
\end{table*}

\subsection{Statistical Significance of Errors}
\label{app:significance}

The errors (recognition blindness and over-sensitivity) observed across the union of the top 10 highest-error target groups for Claude and Mistral were statistically significant. Significance was assessed using a one-sided binomial test under the null hypothesis that the true error rate is zero. For all groups in this set, the p-values were less than 0.05, indicating that the probability of observing such errors by chance alone was below 5\%.

To further assess the reliability of error rates across target groups, we calculated 95\% confidence intervals using the Wilson score interval by \citet{Wilson01061927}. This method was selected over the normal approximation because it provides more accurate bounds, particularly for small sample sizes and when observed proportions are near 0 or 1. See Figures \ref{fig:cm_fn_sig}, \ref{fig:cm_fp_sig}.

\begin{figure}[!htb]
    \centering
    \includesvg[width=1\linewidth]{figures/mistral_claude_false_negatives_comparison_with_ci.svg}
    \caption{Recognition blindness of Claude and Mistral with confidence intervals for each target group}
    \label{fig:cm_fn_sig}
\end{figure}

\begin{figure}[!htb]
    \centering
    \includesvg[width=1\linewidth]{figures/mistral_claude_false_positives_comparison_with_ci.svg}
    \caption{Over-sensitivity of Claude and Mistral with confidence intervals for each target group}
    \label{fig:cm_fp_sig}
\end{figure}

\section{Comparison with Traditional Approaches}
\label{app:traditional_comparison}

For comparison with modern LLM approaches, Table \ref{table:metrics} presents evaluation results for the NJH classifier from \citet{bianchi-etal-2022-just}. This model is a RoBERTa-based classifier fine-tuned to predict eight different labels of uncivil language: Profanity, Insults, Character Assassination, Outrage, Discrimination, Hostility, Incivility, and Intolerance. Following the authors' description, we mapped the `Hostility' label to dehumanization since they stated this category encompasses dehumanizing language. The model's very low performance may be due to the fact that it wasn't specifically trained to identify dehumanization as a distinct phenomenon, but rather as part of a broader `Hostility' category. The stark performance difference highlights the significant advantages of large language models for this task.

\begin{table*}[!ht]
\centering
\begin{tabular}{lc}
\hline
{Metric} & {Value (\%)} \\
\hline
Accuracy & 50.77 \\
Precision & 64.58 \\
Recall & 3.42 \\
F$_1$(other) & 66.59 \\
F$_1$(dehum. (Hostility)) & 6.50 \\
\hline
\end{tabular}
\caption{Evaluation results of NJH classifier introduced by \citet{bianchi-etal-2022-just}}
\label{table:metrics}
\end{table*}

\section{Target Group Frequencies in Evaluation Subsets}
\label{app:target_frequencies}

This section presents the frequency distribution of target groups in our evaluation subsets. These distributions provide important demographic context about the dataset composition and help readers understand the diversity of target groups represented in our analysis.

\subsection{Dehumanization vs. Hate Subset}

Table \ref{tab:dehum_vs_hate_targets} presents the top 15 most frequent target groups in the \texttt{Dehumanization vs. Hate} subset, which consists of dehumanization instances and instances from other hate speech labels.

\begin{table}[!htb]
\centering
\begin{tabular}{lc}
\toprule
{Target Group} & {Frequency} \\
\midrule
Black people & 385 \\
Women & 187 \\
Jewish people & 123 \\
Muslims & 113 \\
Immigrants & 89 \\
Trans people & 70 \\
Refugees & 68 \\
Gay people & 56 \\
People with disabilities & 51 \\
South Asians & 45 \\
African & 37 \\
Working Class & 34 \\
Arabs & 33 \\
Mixed race & 31 \\
Travellers & 29 \\
\bottomrule
\end{tabular}
\caption{Top 15 most frequent target groups in the \texttt{Dehumanization vs. Hate} subset.}
\label{tab:dehum_vs_hate_targets}
\end{table}

\subsection{General Dehumanization Subset}

Table \ref{tab:general_dehum_targets} presents the top 15 most frequent target groups in the \texttt{General Dehumanization} subset, which contains dehumanization instances and a mix of other hate and non-hate labels.

\begin{table}[!htb]
\centering
\begin{tabular}{lc}
\toprule
{Target Group} & {Frequency} \\
\midrule
Black people & 335 \\
Women & 99 \\
Jewish people & 90 \\
Immigrants & 74 \\
Muslims & 68 \\
Refugees & 57 \\
Trans people & 48 \\
Gay people & 39 \\
South Asians & 36 \\
African & 33 \\
Travellers & 25 \\
Arabs & 25 \\
Mixed race & 23 \\
Working Class & 22 \\
People with disabilities & 21 \\
\bottomrule
\end{tabular}
\caption{Top 15 most frequent target groups in the \texttt{General Dehumanization} subset.}
\label{tab:general_dehum_targets}
\end{table}

\section{Fine-grained Hate Type Classification Experiment}
\label{app:fine_grained}

This section provides additional details about the fine-grained hate type classification experiment. In this experiment, models were tasked with classifying texts into specific hate speech types rather than making binary dehumanization judgments.

\subsection{Prompt Template}

For this experiment, we used a few-shot prompting approach with the following template:

\begin{quote}
\small
\texttt{You are an expert in psychology and natural language processing (NLP).\\
Analyze the following text to determine whether it contains hate speech.\\
\\
Step 1: Identify any group(s) targeted in the text.\\
Step 2: Based on your analysis, classify the text using one of the following labels:\\
Not Hate, Derogation, Animosity, Threatening Language, Support for Hateful Entities, Dehumanization\\
\\
- If the text does not contain hate speech, return 'Not Hate'.\\
- If it does, choose the most appropriate hate speech category from the list.\\
\\
Refer to these examples for guidance: [examples provided below]\\
\\
Text: [input text]\\
\\
Return ONLY ONE label from the list above.}
\end{quote}

\subsection{Example Set}

The few-shot examples provided to the models included one instance of each hate speech type, as shown in Table \ref{tab:fine_grained_examples}.

\begin{table*}[!htb]
\centering
\small
\begin{tabular}{p{3cm}p{9cm}}
\toprule
{Hate Type} & {Example} \\
\midrule
Not Hate & Help please! I need new mates willing to go beat up some nasty brothers. \\
\midrule
Derogation & He's uneducated like most <TARGET> travellers. \\
\midrule
Animosity & Fucking delusional to think he's a <TARGET> when I can see his bulge. \\
\midrule
Threatening Language & I agree. We need to work together to kill all the <TARGET> off. \\
\midrule
Support for Hateful Entities &  I wish I was literally in nazi germany in 1937. \\
\midrule
Dehumanization & swine is an apt alternative to naming <TARGET>, they're fucking gross! \\
\bottomrule
\end{tabular}
\caption{Examples used in the fine-grained hate type classification experiment prompt, providing one instance for each type.}
\label{tab:fine_grained_examples}
\end{table*}

\subsection{Performance of Models}

Table \ref{tab:results_model_prompt_perf} provides the metrics for Claude and GPT models under few-shot prompting in the fine-grained hate type classification experiment, revealing that while Claude achieves balanced precision and recall, GPT tends toward higher precision at the expense of recall.

\begin{table*}[!htb]
    \centering
    \begin{tabular}{llcccc} 
    \toprule
    Model & Precision & Recall & F1-score & Accuracy \\ 
    \midrule
    Claude & 85.97 & 70.98 & 77.76 & 64.84\\
    \midrule
    GPT  & 89.31 & 55.21 & 68.24 & 55.52 \\
    \bottomrule
    \end{tabular}
    \caption{Performance comparison of Claude and GPT models.}
    \label{tab:results_model_prompt_perf}
\end{table*}

\end{document}